\documentclass[sigconf, natbib=true]{acmart}
\usepackage{multirow}
\usepackage{tcolorbox}
\usepackage{graphicx}
\AtBeginDocument{%
  \providecommand\BibTeX{{%
    \normalfont B\kern-0.5em{\scshape i\kern-0.25em b}\kern-0.8em\TeX}}}

\setcopyright{acmcopyright}
\copyrightyear{2018}
\acmYear{2018}
\acmDOI{XXXXXXX.XXXXXXX}

\acmConference[Conference acronym 'XX]{Make sure to enter the correct
  conference title from your rights confirmation emai}{June 03--05,
  2018}{Woodstock, NY}
\acmPrice{15.00}
\acmISBN{978-1-4503-XXXX-X/18/06}

\begin{document}

\title{Temporal Data Meets LLM  - \\ Explainable Financial Time Series Forecasting}

\author{Xinli Yu}
\email{xinliy@amazon.com}
\orcid{0009-0000-0374-5946}
\affiliation{%
  \institution{Amazon AWS}
  \city{Seattle}
  \state{WA}
  \country{USA}
}

\author{Zheng Chen}
\email{zgchen@amazon.com}
\orcid{0000-0003-4406-2193}
\affiliation{%
  \institution{Amazon Alexa AI}
  \city{Seattle}
  \state{WA}
  \country{USA}
}

\author{Yuan Ling}
\email{yualing@amazon.com}
\orcid{1234-5678-9012}
\affiliation{%
  \institution{Amazon Benchmarking}
  \city{Seattle}
  \state{WA}
  \country{USA}
}

\author{Shujing Dong}
\email{shujdong@amazon.com}
\orcid{1234-5678-9012}
\affiliation{%
  \institution{Amazon Benchmarking}
  \city{Seattle}
  \state{WA}
  \country{USA}
}

\author{Zongyi Liu}
\email{joeliu@amazon.com}
\orcid{1234-5678-9012}
\affiliation{%
  \institution{Amazon Benchmarking}
  \city{Seattle}
  \state{WA}
  \country{USA}
}

\author{Yanbin Lu}
\email{luyanbin@amazon.com}
\affiliation{%
  \institution{Amazon Alexa AI}
  \city{Seattle}
  \state{WA}
  \country{USA}
}


\begin{abstract}
This paper presents a novel study on harnessing Large Language Models’ (LLMs) outstanding knowledge and reasoning abilities for explainable financial time series forecasting. The application of machine learning models to financial time series comes with several challenges, including the difficulty in cross-sequence reasoning and inference, the hurdle of incorporating multi-modal signals from historical news, financial knowledge graphs, etc., and the issue of interpreting and explaining the model results. In this paper, we focus on NASDAQ-100 stocks, making use of publicly accessible historical stock price data, company metadata, and historical economic/financial news. We conduct experiments to illustrate the potential of LLMs in offering a unified solution to the aforementioned challenges. Our experiments include trying zero-shot/few-shot inference with GPT-4 and instruction-based fine-tuning with a public LLM model Open LLaMA. We demonstrate our approach outperforms a few baselines, including the widely applied classic ARMA-GARCH model and a gradient-boosting tree model. Through the performance comparison results and a few examples, we find LLMs can make a well-thought decision by reasoning over information from both textual news and price time series and extracting insights, leveraging cross-sequence information, and utilizing the inherent knowledge embedded within the LLM. Additionally, we show that a publicly available LLM such as Open-LLaMA, after fine-tuning, can comprehend the instruction to generate explainable forecasts and achieve reasonable performance, albeit relatively inferior in comparison to GPT-4. 

\end{abstract}



\received{20 February 2007}
\received[revised]{12 March 2009}
\received[accepted]{5 June 2009}

\maketitle

\vspace{-0.2cm}

\section{Introduction}

The rapid advancements in Machine Learning (ML) and Artificial Intelligence (AI) technologies over the past few years have opened up numerous opportunities and challenges across various domains, including the realm of financial markets \cite{kovalerchuk2005data, bahrammirzaee2010comparative, qi2018fintech}. In particular, the task of financial time series forecasting, a key element in strategic decision-making and policy formation, has witnessed significant technological innovations, from statistical/econometric time series techniques  \cite{hardle1997review, andersen2009handbook, chen2011review, patton2012review}, to machine learning techniques \cite{kim2003financial, yoo2005machine, krollner2010financial}, to deep learning \cite{dingli2017financial, junior2018generalized, sezer2020financial, leung2021financial, lara2021experimental}. Despite these advancements, there are several inherent challenges associated with the deployment of ML/AI models in finance.

One challenge lies in the realm of \textbf{cross-sequence reasoning and inference}, a vital aspect for understanding temporal patterns and making accurate predictions. The current approaches include time-series correlation analysis \cite{plerou1999universal, gopikrishnan2000scaling, conlon2009cross, chen2018correlated} and clustering \cite{rani2012recent, babu2012clustering, aghabozorgi2015time}. Deep learning has recently been leveraged to learn from the complex latent dependencies among time series \cite{hua2019deep, maulik2020time, song2021capturing, nguyen2021temporal}. Despite these advancements, existing methods have yet to effectively capture the intricate dependencies characteristic of time series data. The varying design, implementation, and data requirements of these methods further creates a barrier for their widespread application in the field.

Another notable hurdle involves handling \textbf{complex multi-modal financial temporal data} that extends beyond numeric sequences. The data may encapsulate diverse sources such as historical news, financial knowledge graphs, social media activities, and various other market indicators. There has been recent effort leveraging \textit{statistical inference} \cite{kanungsukkasem2019financial}, RNN/CNN with text embedding \cite{vargas2017deep}, \textit{graph neural networks} \cite{cheng2022financial}, etc. to integrate the complex information.

Lastly, but of utmost importance, the issue of \textbf{interpretability and explainability} poses significant challenges to the trustworthiness of machine learning and deep learning models. The majority of existing deep learning models operate as black boxes, offering little insight into their decision-making processes. This lack of transparency sometimes raises concerns about the reliability of their results and impedes user trust. This is particularly relevant in sensitive fields like finance, where substantial investments and assets are at stake. There is recent study trying to understand deep-learning based predictions mainly through the attention scores \cite{hsieh2021explainable}, but such insight is still not readily human readable and still requires substantial interpretation effort.

The recent advancement of \textit{Large Language Models} (LLMs) \cite{Brown2020LanguageMA,Touvron2023LLaMAOA,brown2020language,OpenAI2023GPT4TR} potentially lend us a powerful tool to address all above challenges in a unified, flexible way. 

First, \textbf{LLMs can learn the complex relations among sequences}. LLMs are so far the most powerful Transformer-based models, and there has been abundant previous researches showing Transformer-based models are capable of learning the underlying complex relations among textual sequences \cite{yun2019graph, rong2020self, zhang2020graph, dwivedi2020generalization, ying2021transformers} and solving quantitative problems \cite{wei2022chain, lewkowycz2022solving, imani2023mathprompter}. It is reasonable to expect the potential of LLMs understanding complex dependencies among numeric time series augmented by temporal textual sequences. 

Secondly, \textbf{LLMs have demonstrated outstanding reasoning and inference capability over multi-modal data}. By design, LLMs are proficient at learning from a broad spectrum of data sources and types. They are trained on a vast amount of texts from the internet, encompassing a wide range of topics, styles, and formats. This equips them to handle diverse input data, such as numerical, textual, structured data \cite{wu2023bloomberggpt, shen2023efficient}. This multi-modal data handling capability could be particularly useful for financial forecasting, where crucial information often comes from disparate sources, such as numerical market data, textual news articles, and social media posts. 

Lastly, \textbf{LLMs are natural explainers that generate human readable explanations providing insight into a decision}. One of the key advantages of LLMs is their ability to generate natural language text that is coherent, contextual, and comprehensive. This allows them to provide human-readable explanations for their decisions \cite{zhao2023survey}. Furthermore, through Chain-of-Thoughts (COT) or step-by-step thinking \cite{wei2022chain, zhang2023multimodal, lightman2023let}, beyond a few sentences of explanation, LLMs can even generate detailed step-by-step reasoning to reveal the decision-making process. 

The following summarizes the main contributions of this paper,

\vspace{-0.3cm}

\begin{itemize}
    \item This paper takes a novel exploration to study LLMs' potential to the valuable task of explainable financial time series forecasting. For this paper, we focus on the NASDAQ-100 stock price time series. To the best of our knowledge, there is not yet public studies on this topic to date.
    \item We experiment with a combination of zero-shot/few-shot inference techniques with the state-of-the-art AI model GPT-4 \cite{OpenAI2023GPT4TR}, and instruction-based fine-tuning using Open LLaMA \cite{openlm2023openllama}. Our experiment results also show that the technique of chain-of-thoughts helps boost the performance in most of the experiments.
\item We compare our proposed LLM approaches with existing methods, which include an ARMA-GARCH model and a gradient-boosting tree model. We demonstrate even zero-shot inference using GPT-4 can outperform a boosting-tree model with about $\sim$300 features.
\end{itemize}

\section{Related Works}

The field of financial time series forecasting has been a subject of extensive research, with various methodologies being proposed over the years. While traditional statistical methods and machine learning techniques have made significant contributions to this field, the advent of LLMs presents new and significant potentials. 

\subsection{Traditional Statistical/Econometric Methods}

Traditional statistical/econometric methods have long been the cornerstone of financial time series forecasting. Techniques such as ARMA-GARCH models have been widely used due to their ability to capture dependencies and volatility clustering in financial time series \cite{drost1993temporal, francq2004maximum, andersen2009handbook, henneke2011mcmc}. These models have been extended and modified in various ways to better capture the complexities of financial markets \cite{tang2003finite, ghahramani2006financial, hossain2011comparison, ma2013research}. Other popular statistical/econometric methods for financial time series include Vector Autoregressive Models (VAM) \cite{zivot2006vector}, State-Space Models and the Kalman Filter \cite{de1983claims}, Diffusion Models \cite{fan2005selective}, Vector Error Correction Model (VECM) \cite{johansen1995likelihood}, Dynamic Stochastic General Equilibrium (DSGE) \cite{smets2003estimated}, etc.

\subsection{Machine Learning Techniques}

With the advent of machine learning, a variety of models have been applied to financial forecasting. Decision trees, support vector machines, etc., have been actively studied for financial time series prediction \cite{trafalis2000support, yang2002support, pai2005hybrid, wang2006stock, tsai2009stock, li2017comparative}. More recently, deep learning techniques, such as Recurrent Neural Networks (RNNs), Convolutional Neural Networks (CNNs), and Transformer models, have been applied to this task, demonstrating their ability to capture complex, non-linear relationships in the data \cite{dingli2017financial, junior2018generalized, sezer2020financial, leung2021financial, lara2021experimental}.

\subsection{Large Language Models}

The recent development of Large Language Models (LLMs) has opened up new possibilities for financial time series forecasting. LLMs, such as GPT-3 \cite{brown2020language} and GPT-4 \cite{OpenAI2023GPT4TR}, LLaMA\cite{touvron2023llama} (including Alpaca\cite{alpaca}, Vincuna\cite{vicuna2023}), have demonstrated remarkable capabilities in reasoning and understanding complex dependencies in the heterogeneous data, and the ability to generate human-readable explanations for their decisions \cite{zhao2023survey, wei2022chain, zhang2023multimodal, lightman2023let}. However, the application of LLMs in financial time series forecasting with explanation is still a relatively unexplored area, and this paper aims to contribute to this emerging field.

\section{Methodology}

In this study, we focus on the NASDAQ-100 stock price time series, supplemented by metadata about the stock company and relevant financial news data concerning both the specific stock and the broader financial/economic landscape. Our primary focus is on forecasting weekly/monthly \textit{stock returns} (defined as the percentage change in stock price from the beginning to the end of the week/month) with accompanying explanations. This focus aligns well with the expertise of Large Language Models (LLMs). 

We demonstrate our structured design of prompts for LLMs and apply the state-of-the-art GPT-4 model \cite{openai2023gpt4} for zero-shot and few-shot inference tasks. For fine-tuning, we utilize the publicly available Open LLaMA \cite{openlm2023openllama}. We also incorporate the technique of Chain of Thoughts (COT) \cite{wei2022chain, lightman2023let}, which has been found to enhance the effectiveness of LLMs in other research studies. 

\subsection{Data}
\label{sec:data}

\subsubsection{Stock Price Data} We download daily NASDAQ-100 stock\footnote{Using the stock list of Jun 01, 2023} price data from Yahoo Finance using yfinance package (\url{pypi.org/project/yfinance/}). 

In this paper, we first normalize the numeric price time series as a percentage-change time series, and then categorize the percentage change into bins. For example, for weekly forecasting, we categorize price change between this week and last week into 12 bins "D5+", "D5", "D4", "D3", "D2", "D1", "U1", "U2", "U3", "U4", "U5", "U5+", where "D5+" means price dropping more than 5\%, "D$i$" ($i$=5,4,3,2,1) means price dropping between $(i-1)$\% and $i$\%, "U5+" means price rising more than 5\%, "U$i$" ($i$=1,2,3,4,5) means price rising between $(i-1)$\% and $i$\%. The number of bins might vary for inference at different granularity. For example, for monthly inference, we allow $i$ be up to 10, and we have corresponding "D10+" and "U10+" categories.

\subsubsection{Company Profile Data} We use GTP-4 to generate company description, general positive/negative factors that might impact the company's stock price. Figure \ref{fig:company_profile} is an example of the prompt to ask GPT-4 to generate the company profile, and the GPT-4 response.

\begin{figure*}
\begin{tcolorbox}
\textbf{Prompt}: Generate a short description for stock AAPL' s company. Also list general positive and negative factors that might impact the stock price; be brief and use keywords. Consider diverse general factors, such as macro economic situation (e.g. inflation, CPI growth), business factors (e.g. sales, investment, products), technology factors (e.g. innovation), and others. Use format Description: ..., Positive Factors: ..., Negative factors: ... \\

\vspace{0.05cm}

\begin{small}
\textbf{GPT-4 Response}:

\textit{Description}: 

AAPL (Apple Inc.) is a world-renowned technology company based in Cupertino, California. Known for its sleek designs and innovative technologies, the company produces a wide range of products including the iPhone, iPad, Mac, Apple Watch, and services like iCloud, App Store, and more. \\

\vspace{0.01cm}

\textit{Positive Factors}:\\
    Macroeconomic: Economic growth, increased consumer spending\\
    Business: Strong sales, expansion into new markets, positive earnings reports, strategic acquisitions\\
    Technology: Innovation, new product launches, advancements in artificial intelligence and augmented 
    reality\\
    Brand: Strong global brand, loyal customer base\\
    Legal/Regulatory: Favorable policy changes, successful patent applications\\
    
\vspace{0.01cm}

\textit{Negative Factors}:\\
    Macroeconomic: Inflation, economic downturn, trade restrictions\\
    Business: Decreased sales, poor earnings reports, lack of new investments\\
    Technology: Slow pace of innovation, delays in product launches\\
    Brand: Damage to brand reputation, loss of customer loyalty\\
    Legal/Regulatory: Antitrust investigations, unfavorable policy changes, patent disputes\\
    Environmental: Failure to meet sustainability goals, negative environmental impact\\

\end{small}

\end{tcolorbox}
\caption{An example of a stock's company profile consisting of the company description, the general positive/negative factors impacting the company's stock prices.}
\label{fig:company_profile}
\end{figure*}

\subsubsection{Finance/Economy News Data} We use \textit{Google Custom Search API} to obtain stock top-5 news stories on a weekly basis for each of the NASDAQ-100 stocks. After that, we use GPT-4 to generate a summary and extract keywords from each obtained news article. An example of prompt and GPT-4 response is shown in Figure \ref{fig:news_summary_keywords}. 

In addition, a similar method is applied to obtain top-5 news stories about macro economy and finance status of each week.

To reduce input size, We further generate \textit{meta summary and keywords} for each week using GPT-4, given all the top story summaries and keywords of the week. We only use the meta summary and the keywords in this paper's experiments. An example of meta summary and keywords is shown in Figure \ref{fig:news_meta_summary_keywords}; they look similar to the example in Figure \ref{fig:news_summary_keywords}, but much condensed. 

\begin{figure*}
\begin{tcolorbox}

\textbf{Prompt} [template, those inside \{\} are placeholders]:
Please summarize the following noisy but possible news data extracted from web page HTML, and extract keywords of the news. The news text can be very noisy due to it is HTML extraction. Give formatted answer such as Summary: ..., Keywords: ... The news is supposed to be for \{symbol\} stock. You may put 'N/A' if the noisy text does not have relevant information to extract. \\ 

News: \{news\} \\

\vspace{0.05cm}

\begin{small}
\textbf{GPT-4 Response}: \\
\textit{Summary}: Apple Inc is postponing perks for some corporate divisions and expanding a cost-cutting effort, joining its peers in streamlining operations. The company proposes to forego bonuses or promotions for April, with the payments occurring only in October. Apple is also restricting hiring and leaving additional positions open after employee departure. This is triggering fear of downsizing among employees.\\
\textit{Keywords}: Apple, bonus, hiring, cost-cutting, promotions, downsizing, employees
\end{small}

\end{tcolorbox}
\caption{An example of news summary and keywords extracted from one news article for a stock's company (AAPL in this case), generated by GPT-4. The original news is at \url{https://sports.yahoo.com/apple-joins-cost-cut-bandwagon-145845685.html}. The prompt in this example is a template. Texts inside [] are comments and are not really in the prompt when we submit it to the LLM, and we will fill stock symbol and the news into the placeholders enclosed by \{\}.}
\label{fig:news_summary_keywords}
\end{figure*}

\begin{figure*}
\begin{tcolorbox}
\textit{Summary}: Apple Inc.'s stock is displaying rising relative strength, although shy of a key benchmark, and the company has been highlighted as a top-performing stock due to its sales of 1.3 billion iPhones and a significant \$19.4 billion profit last quarter. As the NASDAQ surges with Apple as a big cap, Apple expands cost-cutting measures, postpones perks, restricts hiring, and leaves positions open, instigating downsizing concerns. The overall stock market performance is boosted by a \$30 billion deposit infusion for struggling firms.

\textit{Keywords}: Apple Inc., stock, relative strength, cost-cutting, downsizing, NASDAQ, \$30 billion deposit infusion, iPhones, profit.
\end{tcolorbox}
\caption{An example of one week's meta summary and keywords condensed from all the company's summaries and keywords from the week.}
\label{fig:news_meta_summary_keywords}
\end{figure*}

\subsection{Instruction-Based Zero-shot/Few-shot Inference with LLMs}

In zero-shot and few-shot inference, LLMs demonstrate their ability to generate responses either without any additional examples (zero-shot) or based on a minimal number of examples beyond the original training set (few-shot).

In our zero-shot/few-shot inference experiment, we utilize an instruction-based prompt. The structure of our prompt, illustrated in Figure \ref{fig:inference_prompt}, includes instructions, the company profile, a historical temporal news summary/keywords sequence intermixed with the categorized stock price time series, and \textit{cross-sequence few-shot learning examples}.

To avoid unnecessary repetition in the prompt text, we purposely provide few-shot learning examples from stocks similar to the subject of interest. This design also assists us in demonstrating that the LLM can consider cross-sequence information from various stocks. To identify similar stocks, we query GPT-4 with a question such as "List the top 3 NASDAQ stocks most similar to AAPL". A typical response, such as "MSFT, GOOGL, AMZN", showcases the LLM's comprehension of the relationships between various financial entities and concepts. In employing an LLM, we are implicitly leveraging its extensive knowledge of financial entities and concepts.

The prompt structure and instructions have been empirically tweaked. For instance, we divided the instruction into two parts, positioning them at the beginning and end of the prompt, which aids the model in better recognizing its task: to predict next week's summary and keywords, rather than summarizing historical data. The predicted summary and keywords serve as the explanation for the corresponding stock return prediction.

We also experimented with the Chain-of-Thoughts approach \cite{wei2022chain, zhang2023multimodal, lightman2023let}, i.e., the idea of "step-by-step thinking", by appending the instruction "Can you reason step by step before finalizing the output?" to the end of the prompt. To our surprise, this notably improved the performance by a few points (see Section \ref{sec:performance_evaluation}). The result of the step-by-step thinking process in response to Figure \ref{fig:inference_prompt} is illustrated in Figure \ref{fig:step_by_step_thinking}, where it is evident that GPT-4 identifies a previously overlooked crucial point about "earnings reports" when explicit reasoning steps are generated.

\begin{figure*}
\begin{tcolorbox}

\textbf{Prompt} [template, those inside \{\} are placeholders]:

\textit{Instruction}:
Forecast next week stock return (price change) for {symbol}, given the company profile, historical weekly news summary, keywords, and stock returns, and optionally the examples from other stocks of a similar company. \\

The trend is represented by bins "D5+", "D5", "D4", "D3", "D2", "D1", "U1", "U2", "U3", "U4", "U5", "U5+", where "D5+" means price dropping more than 5\%, D5 means price dropping between 4\% and 5\%, "D4" means price dropping between 3\% and 4\%, "U5+" means price rising more than 5\%, "U5" means price rising between 4\% and 5\%, "D4" means price rising between 3\% and 4\%, etc.

\textit{Company Profile}: \{company\_profile\} [company description, general positive/negative factors as shown in Figure \ref{fig:company_profile}.]

\textit{Recent News}: News are ordered from oldest news to latest news.

8 Weeks ago. 

\{meta\_news\_summary\_company\} [meta news summary and keywords of the week for the company] 

\{meta\_news\_summary\_macro\} [meta news summary and keywords for the macro economy and finance status of the week] 

... 

7 weeks ago. 

...

Last week. 

... 

\textit{Forecasting Examples}:
\{few\_shot\_learning\_examples\_from\_similar\_stocks\}
[We use forecasting examples from top-3 similar stocks as the few-short learning examples. The example is structured similar to this prompt, with company profile and recent news, but without the instruction texts.] 
\\

Now predict what could be the next week's Summary, Keywords, and forecast the Stock Return. The predicted Summary/Keywords should explain the stock return forecasting. You should predict what could happen next week. Do not just summarize the history. The next week stock return need not be the same as the previous week. Use format Summary: ..., Keywords: ..., Stock Return: ... \\

\vspace{0.05cm}

\begin{small}
\textbf{GPT-4 Response} [Forecasting week 05/01/2023 to 05/07/2023]: 

\textit{Summary}: Apple's stock is expected to see fluctuations in the week to come against a backdrop of mixed news. The company's announced expansion into India, coupled with the introduction of high-yield savings accounts, initially provided a boost to the stock. However, growing concerns surrounding reduced PC shipments and apprehensions about the company's upcoming earnings injected a degree of uncertainty. Despite these challenges, Apple's unwavering focus on cost control and the anticipation surrounding innovative offerings, such as the forthcoming mixed-reality headset, shored up investor confidence. Consequently, the stock maintained a general upward trend throughout the week, albeit punctuated by a few dips. The broader economic climate on Wall Street also had a role to play, with a recovering sentiment in Q1 towards high-tech companies like Microsoft and Alphabet. These companies released pivotal financial reports and ground-breaking products during the period, influencing the tech sector's performance, potentially impacting Apple's stock trajectory. \\

\textit{Keywords}: AAPL, mixed-reality headset, India expansion, high-yield savings account, cost control strategies, Bernstein, Wall Street, economic recovery, tech sector performance \\

\textit{Stock Return}: U1 \\

\vspace{0.05cm}

\textbf{Ground Truth} [Summary from actual news from 05/01/2023 to 05/07/2023 and the actual return]: 

\textit{Summary}: Apple's stock saw significant gains after outperforming Wall Street's fiscal second quarter targets, driven by robust iPhone sales and strong services, despite posting a decrease in revenue for the second consecutive quarter and facing potential competition in the VR headset market from Reliance Jio's JioDive. The company also noted an improvement in supply and strength in emerging markets, further boosting investor confidence. \\

\textit{Keywords}: Apple, stock, Wall Street, fiscal second quarter, iPhone, services sales, quarterly earnings, revenue, declining revenue, emerging markets, improved supply, Virtual Reality, competition, JioDive, Reliance Jio. \\

\textit{Stock Return}: U3
\end{small}

\end{tcolorbox}
\caption{The prompt structure for the experiments in this paper with LLMs. We also give an example of GPT-4 response to a concrete prompt constructed from information on and before 04/30/2023. We notice the cross-sequence information and macro-economy information are obviously considered in the LLM's reasoning. The stock return forecast U1 is for the next week from 05/01/2023 to 05/07/2023. The actual market performance for AAPL is U3. We note Apple 2023 Q2 earning call happened on May 04 beat expectation, which maybe the major contributor to the higher-than-forecasting gain in the week.}
\label{fig:inference_prompt}
\end{figure*}

\begin{figure*}
\begin{tcolorbox}

\textbf{Prompt}:
\{The same prompt as in Figure \ref{fig:inference_prompt}.\} \\
Can you reason step by step before the finalized output? \\

\vspace{0.05cm}

\begin{small}
\textbf{GPT-4 Response} [Forecasting week 05/01/2023 to 05/07/2023]: \\
Given the information provided, we can see that Apple (AAPL) has had a number of positive developments and a few potential concerns in the recent past.

Positive Developments:
\begin{itemize}
    \item Recent sales and earnings have been strong, and the company appears to be exploring new markets such as India.
    \item The upcoming mixed-reality headset and the "Apple Pay Later" service show that Apple is still innovating and pushing into new product areas.
    \item The cost-cutting measures and the plan to use 100\% recycled cobalt show that Apple is taking steps to improve its efficiency and sustainability, which could be attractive to certain investors.
    \item Winning the lawsuit against Britain's antitrust regulator and avoiding potential penalties.
    \item AAPL stock has been trending upwards recently, with multiple weeks of upward movement (U5, U4, U3, U2), and only a single day of slight downward movement (D1).
\end{itemize}

Potential Concerns:
\begin{itemize}
    \item Apple is cutting costs, delaying bonuses, and slowing down hiring in certain areas, which could potentially affect employee morale and productivity.
    \item The negative sentiment from analyst Toni Sacconaghi about the overestimation of Q3 and Q4 revenues by Wall Street could potentially impact the stock price. \\
\end{itemize}

In addition to above factors, the general direction of the stock market appears to be on an upward trajectory, as evidenced by the recent increases in the Dow Jones, S\&P 500, and Nasdaq. In recent weeks, the stock performances of high-tech giants such as Microsoft and Alphabet have also shown considerable strength. Furthermore, recent positive developments in the macroeconomic environment, including a surge in consumer spending, seem encouraging. \\

Now, we can generate a prediction for next week's Summary, Keywords, and Stock Return. However, it's important to note that while the information provided is used as a basis, this is a prediction and the actual outcome may be different. \\

\vspace{0.05cm}

\textit{Summary}: Apple's mixed-reality headset previews have received positive feedback from industry experts, increasing investor confidence. However, apprehension about cost-cutting measures and fears of slower hiring might add some uncertainty. Wall Street anticipates a strong earnings report, boosting stock morale. Furthermore, Apple's expansion into India continues to show promise.  \\

\textit{Keywords}: Apple, mixed-reality headset, investor confidence, cost-cutting, hiring, Wall Street, earnings report, expansion, India. \\

\textit{Stock Return}: U2. Given the positive trends and promising developments, the stock is likely to experience upward movement. However, cost-cutting concerns may moderate gains, leading to a prediction of "U2", suggesting a rise of 1\%~2\% next week.
\end{small}

\end{tcolorbox}
\caption{GPT-4 outputs its detailed reasoning steps if we simply add an instruction "Can you reason step by step before the finalized output?" to the end of the prompt in Figure \ref{fig:inference_prompt}. With detailed reasoning steps, GPT-4 captures a previously missed point "Wall Street anticipates a strong earnings report, boosting stock morale", and amends the stock return forecast as U2.}
\label{fig:step_by_step_thinking}
\end{figure*}

\subsection{Instruction-based Fine-tuning with Open LLaMA}

We perform instruction-based fine-tuning using Open LLaMA 13B model to see how well a publicly available model could perform in comparison to GPT-4, especially after fine-tuning. The Open LLaMA 13B model, in its zero-shot inference, typically tends to replicate portions of the prompt rather than executing the instructions in the prompt effectively. Therefore, it is incapable of properly handling instruction-based prompts as shown in Figure \ref{fig:inference_prompt} without undergoing a process of fine-tuning. Therefore we focus on fine-tuning with the Open LLaMA model in this paper.

Instruction-based fine-tuning has been recently shown to be effective in guiding the model's training process with specific directives \cite{alpaca, peng2023instruction}. We created a dataset of 30K weekly forecasting plus 7K monthly forecasting, derived from 5-year historical data spanning from Jun 2017 to June 2022. Unlike GPT-4 that supports up to 8K token size, we need to compress the prompt into 1K tokens for fine-tuning Open LLaMA, due to model and hardware constraints. For each fine-tuning example, we employ GPT-4 to condense the full historical meta news summary/keywords (e.g. from week 8 to the last week as shown in Figure \ref{fig:inference_prompt}) into a single, even more concise summary/keywords pair. Simultaneously, the "Company Profile" and "Forecasting Examples" sections of the prompt are also respectively condensed into more succinct summary paragraphs. 

While it would be ideal for Open LLaMA to manage its own end-to-end experiment, including the task of prompt compression for fine-tuning, we still resort to using GPT-4 right now. This is due to Open LLaMA 13B model's zero-shot summarization capability is considerably inferior to those of GPT-4 in practice. The summaries and keywords extracted by Open LLaMA 13B model often fall short of usability.

Once fine-tuned, the Open LLaMA 13B model demonstrates a much more satisfactory comprehension of the instruction, resulting in the generation of a forecast and an accompanying explanation that appears coherent. This is illustrated in Figure \ref{fig:open_llama_result}. As per the result in section \ref{sec:performance_evaluation}, when it comes to binary classification, the Open LLaMA model's performance is competitive compared to GPT-4. However, we've noticed that the Open LLaMA model has a tendency to produce more extreme predictions, such as U5+ or D5+, which result in a relatively higher squared error.

\begin{figure*}
\begin{tcolorbox}

\textit{Summary}: The stock market is likely to react favorably to Apple's continued focus on cost optimization and revenue generation through new product launches and market expansions. Despite facing regulatory challenges, the company's efforts towards environmental responsibility and diversification of financial offerings are likely to have a positive impact on its long-term prospects. As a result, we expect Apple's stock to see another strong week of growth, with a projected return of U5+.

\textit{Keywords}: Cost-Cutting, New Product Launches, Expansion, Regulation, Supply Chain

\textit{Stock Return}: U5+ 

\end{tcolorbox}
\caption{Explainable forecasting by fine-tuned Open LLaMA 13B model.}
\label{fig:open_llama_result}
\end{figure*}

\section{Experiments}

\subsection{Experiment Setup}

\subsubsection{Data Time Window} The details of the data used in the experiments is as described in Section \ref{sec:data}. We focus on NASDAQ-100 stock return forecasting for this paper. 
\begin{itemize}
    \item The \textit{training/fine-tuning time window} contains 5-year data from Jun 12 2017 to Jun 05 2022. The data in this time window is used for training of the baseline models, and the fine-tuning of the Open LLaMA 13B model.
    \item The \textit{evaluation time window} has 52 weeks spanning from Jun 06 2022 to Jun 04 2023. The evaluation of baseline models, the zero/few-shot inference experiments with GPT-4, and the evaluation of fine-tuned Open LLaMA 13B model, are based on data in this time window.
\end{itemize}

\subsubsection{Baseline Models}
To evaluate the performance of our approach, we include a heuristic baseline using the most-frequent historical bin (i.e. the most frequent bin from historical weeks before the target week to forecast) as the prediction, an ARMA-GARCH model ($p=q=1$) \cite{tang2003finite, ma2013research}, and a gradient-boosting tree model \cite{natekin2013gradient} implemented by LightGBM package \cite{ke2017lightgbm}. These baseline models are trained on the training/fine-tuning data time window, and evaluated on the evaluation time window.

For the gradient-boosting tree model, we include the following features. There are total about 300 features for the tree.
\begin{enumerate}
    \item Historical price time series available in the daily stock price data, including open, close, min, max prices, and the daily trading volume.
    \item The average, medium, min, max, and stddev of a rolling window of size 2, 5, 10, 30, 60, 90 for the above time series.
    \item The stock sector information, and the stock historical earning obtained from Alpha Vantage  (\url{https://www.alphavantage.co/documentation/}).
\end{enumerate}

\subsubsection{Evaluation Metrics} We perform weekly and monthly stock return forecasting with the baselines and LLM-based methods. We treat 4 weeks as one month for convenience, and therefore there are 13 "month"s in the 52-week evaluation time window. 

To evaluate the performance of our forecasting models, we employ three metrics. 
\begin{itemize}
    \item Binary precision assesses the model's ability to correctly predict the general direction of stock price movement, i.e., "Up" (U) or "Down" (D). 
    \item Bin precision, on the other hand, evaluates the model's accuracy in predicting the exact bin from a full list of bins such as "D5+", "D5", "D4", ..., "D1", "U1", ..., "U5", "U5+". 
    \item The MSE of consecutive bin ordinals (e.g., -6 for "D5+", -5 for "D5", ..., 0 for "U1", ..., 4 for "U5", 5 for "U5+") is used to measure the average squared differences between the model's predictions and the actual values. This metric helps to understand the model's tendency to make extreme forecasts when its predictions are incorrect.
\end{itemize}

To evaluate the quality of the forecasting explanation (the predicted next-week/month summary/keywords), we employ ROGUE-1 and ROGUE-2 scores to compare with the actual summary/keywords by GPT-4 extracted from the actual top news of the next week/month. 

\subsection{Performance Evaluation}
\label{sec:performance_evaluation}

The results of our experiments are summarized in Table \ref{tab:performance_stock_price_forecast} and \ref{performance_explanation_quality}. Table \ref{tab:performance_stock_price_forecast} provides a comparative analysis of our LLM-based methods and the baseline models in terms of their performance in forecasting stock returns. Table \ref{performance_explanation_quality}, on the other hand, evaluates the quality of the explanations generated by the LLMs. 

In summary, our results show the effectiveness of LLMs in financial time series forecasting, with "GPT-4 few-shot with COT" consistently showing the best performance in both prediction accuracy and explanation quality. The results also highlight the technique of Chain-of-Thoughts (COT) consistently boosts performance, and the potential of instruction-based fine-tuning with publicly available LLMs like Open LLaMA to achieve reasonable performance in comparison to GPT-4 through fine-tuning with COT.

\subsubsection{Stock Price Forecasting}

From the results of Table \ref{tab:performance_stock_price_forecast}, we observe that both GPT-4 and Open LLaMA 13B model outperform the ARMA-GARCH model and the gradient-boosting tree model in terms of both binary and bin precision. GPT-4, in particular, shows superior performance in both zero-shot and few-shot settings, with the few-shot setting with COT achieving the best performance. In terms of MSE, "GPT-4 few-shot with COT" also achieves the lowest error, indicating that it not only best predicts the direction of the price change but also provides a more accurate estimate of the magnitude of the change.

Open LLaMA 13B model, after fine-tuning, shows competitive performance compared to GPT-4 in terms of binary precision. However, its bin precision is obviously worse, indicating it lacks competitive fine-grained reasoning capability to pick the right bin. It also tends to produce more extreme predictions, resulting in a relatively higher MSE.

\subsubsection{Explanation Quality}

Table \ref{performance_explanation_quality} shows the quality of the explanations generated by the LLMs (GPT-4 and fine-tuned Open LLaMA), evaluated using ROUGE-1 and ROUGE-2 scores for both the summary (S) and keywords (K) of the news.

Again, the results show that "GPT-4 few-shot with COT" achieves the highest ROUGE scores, indicating that it generates the most relevant and accurate explanations for the predictions. Open LLaMA, after fine-tuning with COT, also shows reasonable explanation quality in parallel with GPT-4 results without COT.

\begin{table*}[]
\centering
\resizebox{\textwidth}{!}{%
\begin{tabular}{|l|ccc|ccc|}
\hline
\multicolumn{1}{|l|}{}                & \multicolumn{3}{c|}{\textbf{Weekly}}                                                                    & \multicolumn{3}{c|}{\textbf{Monthly (Every 4   Weeks)}}                                                 \\ \hline
\textbf{}                             & \multicolumn{1}{c|}{\textbf{Binary Precision}} & \multicolumn{1}{c|}{\textbf{Bin Precision}} & \textbf{MSE} & \multicolumn{1}{c|}{\textbf{Binary Precision}} & \multicolumn{1}{c|}{\textbf{Bin Precision}} & \textbf{MSE} \\ \hline
\textbf{Most-Frequent Historical Bin}                  & \multicolumn{1}{c|}{50.7\%}                    & \multicolumn{1}{c|}{16.4\%}             & 43.5         & \multicolumn{1}{c|}{51.4\%}                    & \multicolumn{1}{c|}{17.2\%}             & 155.1        \\ \hline
\textbf{ARMA-GARCH}                   & \multicolumn{1}{c|}{52.4\%}                    & \multicolumn{1}{c|}{11.1\%}             & 22.1         & \multicolumn{1}{c|}{50.5\%}                    & \multicolumn{1}{c|}{6.2\%}              & 90.1         \\ \hline
\textbf{Gradient Boosting Tree Model}                     & \multicolumn{1}{c|}{60.8\%}                    & \multicolumn{1}{c|}{26.4\%}             & 24.3         & \multicolumn{1}{c|}{56.4\%}                    & \multicolumn{1}{c|}{17.7\%}             & 85.6         \\ \hline
\textbf{GPT-4 Zero-Shot}              & \multicolumn{1}{c|}{64.5\%}                    & \multicolumn{1}{c|}{31.2\%}             & 20.5         & \multicolumn{1}{c|}{64.8\%}                    & \multicolumn{1}{c|}{26.0\%}             & 60.1         \\ \hline
\textbf{GPT-4 Few-Shot}               & \multicolumn{1}{c|}{65.8\%}                    & \multicolumn{1}{c|}{32.7\%}             & 20.6         & \multicolumn{1}{c|}{65.3\%}                    & \multicolumn{1}{c|}{26.5\%}             & 58.2         \\ \hline
\textbf{GPT-4 Few-Shot w/ COT}        & \multicolumn{1}{c|}{\textbf{66.5}\%}                    & \multicolumn{1}{c|}{\textbf{35.2}\%}             & \textbf{18.7}         & \multicolumn{1}{c|}{\textbf{69.5}\%}                    & \multicolumn{1}{c|}{\textbf{28.6}\%}             & \textbf{50.4}         \\ \hline
\textbf{Open LLaMA (13B) Fine-Tuned}        & \multicolumn{1}{c|}{62.2\%}                    & \multicolumn{1}{c|}{26.5\%}             & 23.3         & \multicolumn{1}{c|}{60.1\%}                    & \multicolumn{1}{c|}{22.6\%}             & 63.3         \\ \hline
\textbf{Open LLaMA (13B) Fine-Tuned w/ COT} & \multicolumn{1}{c|}{64.7\%}                    & \multicolumn{1}{c|}{30.7\%}             & 21.0         & \multicolumn{1}{c|}{62.2\%}                    & \multicolumn{1}{c|}{24.4\%}             & 63.5         \\ \hline
\end{tabular}%
}
\caption{Performance comparison between the baseline models and LLMs for stock price weekly/monthly forecasting.}
\label{tab:performance_stock_price_forecast}
\end{table*}

\begin{table*}[]
\centering
\resizebox{\textwidth}{!}{%
\begin{tabular}{|l|cccc|cccc|}
\hline
\multicolumn{1}{|l|}{}                & \multicolumn{4}{c|}{\textbf{Weekly}}                                                                                                                     & \multicolumn{4}{c|}{\textbf{Monthly (Every 4   Weeks)}}                                                                                                                       \\ \hline
\textbf{}                             & \multicolumn{1}{c|}{\textbf{ROUGE-1 (S)}} & \multicolumn{1}{c|}{\textbf{ROUGE-2 (S)}} & \multicolumn{1}{c|}{\textbf{ROUGE-1 (K)}} & \textbf{ROUGE-2 (K)} & \multicolumn{1}{c|}{\textbf{ROUGE-1 (S)}} & \multicolumn{1}{c|}{\textbf{ROUGE-2 (S)}} & \multicolumn{1}{c|}{\textbf{ROUGE-1 (K)}} & \multicolumn{1}{c|}{\textbf{ROUGE-2 (K)}} \\ \hline
\textbf{GPT-4 Zero-Shot}              & \multicolumn{1}{c|}{0.2212}               & \multicolumn{1}{c|}{\textbf{0.0675}}               & \multicolumn{1}{c|}{0.1295}               & 0.0447               & \multicolumn{1}{c|}{0.2528}               & \multicolumn{1}{c|}{0.0665}               & \multicolumn{1}{c|}{0.1335}               & 0.0657                                    \\ \hline
\textbf{GPT-4 Few-Shot}               & \multicolumn{1}{c|}{0.2242}               & \multicolumn{1}{c|}{0.0526}               & \multicolumn{1}{c|}{0.1304}               & 0.0454               & \multicolumn{1}{c|}{0.2450}               & \multicolumn{1}{c|}{0.0634}               & \multicolumn{1}{c|}{0.1348}               & 0.0644                                    \\ \hline
\textbf{GPT-4 Few-Shot w/ COT}        & \multicolumn{1}{c|}{\textbf{0.2414}}               & \multicolumn{1}{c|}{0.0543}               & \multicolumn{1}{c|}{\textbf{0.2083}}               & \textbf{0.0869}               & \multicolumn{1}{c|}{\textbf{0.2645}}               & \multicolumn{1}{c|}{\textbf{0.0758}}               & \multicolumn{1}{c|}{\textbf{0.2450}}               & \textbf{0.1025}                                    \\ \hline
\textbf{Open LLaMA (13B) Fine-Tuned}        & \multicolumn{1}{c|}{0.2053}               & \multicolumn{1}{c|}{0.0395}               & \multicolumn{1}{c|}{0.0927}               & 0.0324               & \multicolumn{1}{c|}{0.2242}               & \multicolumn{1}{c|}{0.0474}               & \multicolumn{1}{c|}{0.1167}               & 0.0520                                    \\ \hline
\textbf{Open LLaMA (13B) Fine-Tuned w/ COT} & \multicolumn{1}{c|}{0.2371}               & \multicolumn{1}{c|}{0.0434}               & \multicolumn{1}{c|}{0.1123}               & 0.0425               & \multicolumn{1}{c|}{0.2436}               & \multicolumn{1}{c|}{0.0536}               & \multicolumn{1}{c|}{0.1356}               & 0.0834                                    \\ \hline
\end{tabular}%
}
\caption{Explanation quality evaluation using ROGUE scores, using the GPT-4 summary/keyword extraction of each week's true top news from google search as the ground truth.}
\label{performance_explanation_quality}
\end{table*}

\section{Conclusion}

In this paper, we explored the application of Large Language Models (LLMs) in the field of financial time series forecasting, with a particular focus on NASDAQ-100 stock return prediction. We demonstrated how LLMs, specifically GPT-4 and Open LLaMA, can be utilized to generate forecasts and provide human-readable explanations for their predictions. Our approach involved the use of structured prompts, which included company profile data, historical stock price data, and financial news data, to guide the LLMs in their forecasting tasks.

Our experimental results revealed that LLMs could surpass traditional statistical models and machine learning techniques, such as the ARMA-GARCH model and a gradient-boosting tree model, in performance. Notably, the integration of a step-by-step reasoning process, inspired by the Chain of Thought (COT) approach, significantly enhanced the performance of our LLM-based models. Moreover, our fine-tuning experiments with Open LLaMA demonstrated the feasibility of effectively tuning publicly available LLMs for this task, thereby addressing the inherent challenges of cross-sequence reasoning, multi-modal signals integration, and result interpretability in the financial sector.

In conclusion, our preliminary exploration into the application of LLMs in financial forecasting yielded promising initial results. Despite being in the early stages, the encouraging outcomes provide a strong motivation for further exploration in this direction. As we continue to expand the application of LLMs in the financial domain, we envision a future where financial forecasting is not only more precise but also more comprehensible and transparent. This development could significantly contribute to the transformation of financial decision-making across the entire sector.

Future research will dive deeper into these methodologies, including but not limited to: 1) extending the studies to include more stock indexes such as SP500 and Russell 2000, 2) integrating the research with more data types, such as the macro economy time series, stock trading volumes, and social network data, and 3) exploring the fine-tuning of larger publicly available models, such as a 30B model, to enhance reasoning capabilities.

\bibliographystyle{ACM-Reference-Format}
\bibliography{custom}

\appendix

\end{document}